\documentclass[sigconf, nonacm]{acmart}
\usepackage{float}
\usepackage{subcaption}
\usepackage{tabularray}
\UseTblrLibrary{booktabs} 
\usepackage{color}
\usepackage[ruled,vlined]{algorithm2e}
\usepackage{xspace}
\usepackage{listings}

\usepackage{nicefrac}
\usepackage{graphicx}
\usepackage{textcomp}
\def\BibTeX{{\rm B\kern-.05em{\sc i\kern-.025em b}\kern-.08em
    T\kern-.1667em\lower.7ex\hbox{E}\kern-.125emX}}
\usepackage{booktabs}
\usepackage{multicol}               
\usepackage{multirow}
\usepackage{environ}     
\usepackage[most]{tcolorbox}
\usepackage{tikz}
\usepackage{adjustbox}
\usepackage{xcolor}
\usepackage{cleveref}
\usepackage{caption} 
\usepackage{cuted}    

\definecolor{codegreen}{rgb}{0,0.6,0}
\definecolor{codegray}{rgb}{0.5,0.5,0.5}
\definecolor{codepurple}{rgb}{0.58,0,0.82}
\definecolor{backcolour}{rgb}{0.95,0.95,0.92}

\lstdefinestyle{mystyle}{
    backgroundcolor=\color{backcolour},   
    commentstyle=\color{codegreen},
    keywordstyle=\color{magenta},
    numberstyle=\tiny\color{codegray},
    stringstyle=\color{codepurple},
    basicstyle=\ttfamily\footnotesize,
    breakatwhitespace=false,         
    breaklines=true,                 
    captionpos=b,                    
    keepspaces=true,                 
    numbers=left,                    
    numbersep=5pt,                  
    showspaces=false,                
    showstringspaces=false,
    showtabs=false,                  
    tabsize=2
}

\NewEnviron{wwnn}{{\textcolor{purple}{ \BODY}}}

\newcommand{\CLUB}{CluB\xspace}

\newcommand*\firstpagestarfootnote[1]{%
  \begingroup
    \renewcommand\thefootnote{\fnsymbol{footnote}}
    \setcounter{footnote}{0}                      
    \footnotetext[1]{#1}
  \endgroup}

\newcommand{\method}{BLaST}
\newcommand{\png}{P\&G}

\usepackage{makecell}

\begin{document}

\title{\method{}: High Performance Inference and Pretraining using \linebreak BLock Sparse Transformers}
\fancyhead[L]{\method{}: High Performance Inference and Pretraining using BLock Sparse Transformers}

\author{Patrik Okanovic\textsuperscript{1*}, 
Sameer Deshmukh\textsuperscript{2*}, 
Grzegorz Kwasniewski\textsuperscript{1*},\\
Yi Zhu\textsuperscript{1}, 
Haruto Fujii\textsuperscript{2}, 
Sakina Fatima\textsuperscript{2}, 
Maciej Besta\textsuperscript{1}, 
Kentaro Katayama\textsuperscript{2}, \\
Takumi Honda\textsuperscript{2}, 
Yusuke Nagasaka\textsuperscript{2}, 
Torsten Hoefler\textsuperscript{1}}
\affiliation{%
  \institution{\textsuperscript{1}ETH Zurich, \textsuperscript{2}Fujitsu Pvt. Ltd.}
  \country{\texttt{\{pokanovic,gkwaniewski\}@ethz.ch, deshmukh.sameer@fujitsu.com}}}
\fancyhead[R]{}

\begin{abstract}
The energy consumption of large-scale ML models is dominated by data movement, shuffling billions of parameters across memory hierarchies and data centers. Sparsification offers a principled way to mitigate these costs by pruning redundant weights and activations, thereby reducing data movement. Effective sparsification to prune redundant parameters is still challenging: existing methods incur significant accuracy degradation, performance overhead, or both. We introduce (Bl)ock (a)nd (S)parse (T)ransformers (\method{}), a general, robust, and reliable method for sparsification, applicable to linear layers in all settings. Our method iteratively sparsifies weight matrices into a block sparsity pattern suitable for efficient sparse matrix-matrix (SpMM) multiplication.
\method{} achieves up to 95\% sparsity in MLP weights with negligible accuracy loss (majority $<$2.25\%). We show a 2.2x inference speedup for Llama 3.2 with 16 GPUs, and up to 4.45x reduction in inference memory footprint resulting in a 2.9x reduction in GPU setup and operating costs.

\end{abstract}

\maketitle

\firstpagestarfootnote{Equal contribution}

\section{Introduction}
\label{sec:introduction}
Large Language Models (LLMs) based on the Transformer architecture have driven a dramatic expansion in the adoption of AI across everyday applications. State-of-the-art models~\cite{Geminiteam2024, Touvron2023a, Deepseekai2025, Radford2018, Radford2019, Brown2020} rely on large-scale data centers for both pretraining on massive datasets, and subsequent deployment to serve millions of users. While recent algorithmic innovations have reduced model size and computational demands~\cite{Ainslie2023, Meng2025}, the trend toward ever-larger parameter counts continues unabated~\cite{Shen2025}. Training such models to high accuracy and serving them with low latency requires substantial investment in hardware infrastructure, making cost reduction a central challenge.
These hardware investments are overwhelmingly tailored to Transformer-based architectures~\cite{Vaswani2023}. The associated costs scale with the size of the model's weight matrices: larger models consume more memory, and require more GPUs for storage and computation.

\begin{figure}
    \centering
    \includegraphics[width=\linewidth]{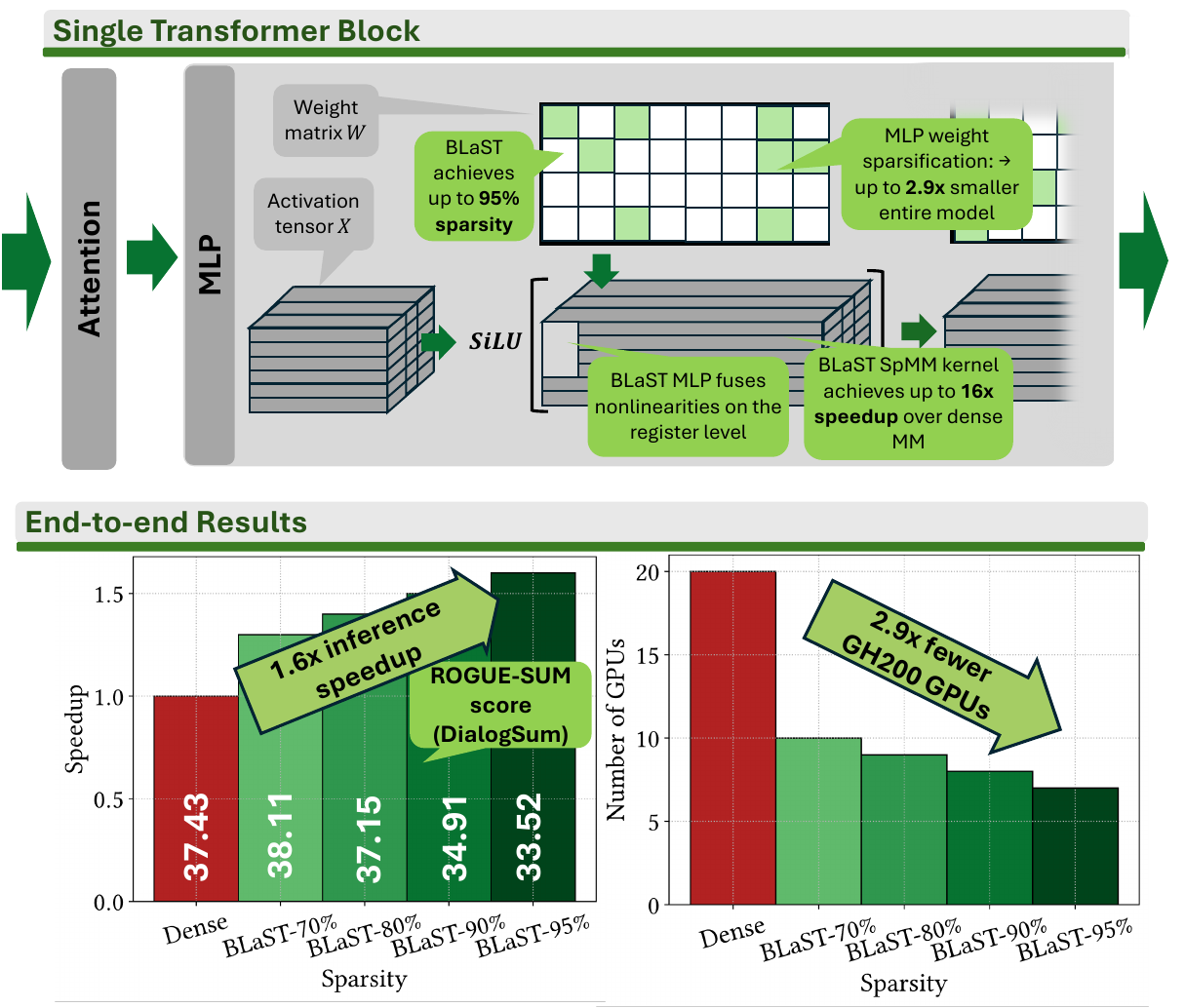}
    \caption{\method{} sparsifies MLP weights up to 95\% with minimal accuracy loss and speeds up end-to-end Llama~3.2--1B inference by up to 1.6$\times$. For Llama~3.2--405B, sparsity reduces required GPUs by up to 2.9$\times$.}
    \label{fig:intro-figure}
\end{figure}

\begin{figure*}
    \centering
    \includegraphics[width=0.9\linewidth]{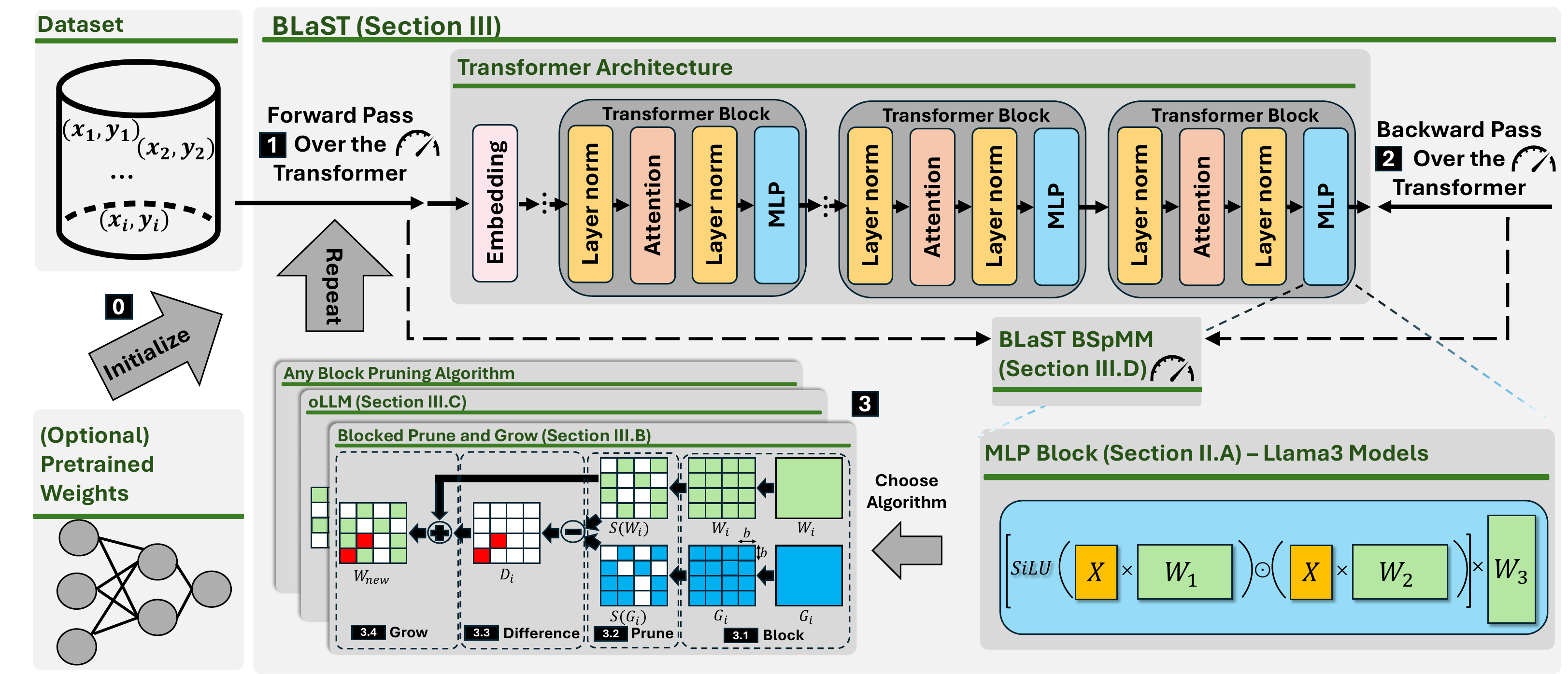}
    \caption{Overview of \method{}. Training or fine-tuning alternates dense updates with a pluggable block-pruning stage (instantiated with Blocked Prune-and-Grow (\Cref{sec:blocked-prune-and-grow}) and oLLM (\Cref{sec:ollm})) while using a fused, high-performance sparse MLP kernel (\Cref{sec:kernel_section}) until the target sparsity is reached.}
    \label{fig:blast_overview}
\end{figure*}

Model pruning~\cite{Hoefler2021,Koch2007,Schotthofer2022,Ceruti2022,Ceruti2022a,Hu2024,Hu2024a,Hubara2021,Zhou2021,Zhou2021a,Evci2021,Frankle2019,Child2019,Pan2024,Mishra2021,Ivanov2023,Hu2021,Buyukakyuz2024,Chen2024,castro2023venom,Frantar2023,Ashkboos2024,Chen2021a,Chen2021,Chen2020,You2022}, i.e. removing redundant parameters, is a principled strategy that simultaneously reduces memory footprint, minimizes data movement, and increases inference throughput. As such, it represents a compelling approach for improving efficiency across training, deployment, and serving of LLMs at scale.
Despite active research, developing a general-purpose sparsification method that simultaneously delivers high sparsity, high performance, and minimal accuracy loss remains difficult. The primary challenge lies in hardware constraints: modern accelerators rely on dense, batched memory access patterns and computation.
Unstructured pruning can reach high sparsity by removing individual weights, but irregular patterns are hard to accelerate on current hardware~\cite{Frankle2019,prasanna2020bert,gordon2020compressing,Chen2020,guo1909reweighted,sanh2020movement}.
Attempts to enforce structured sparsity (e.g., pruning blocks or entire channels) offer better hardware compatibility but typically fail to exceed sufficient sparsity for speedups without incurring significant accuracy loss~\cite{Ashkboos2024}. Even with NVIDIA GPU support~\cite{Mishra2021}, N:M sparsity prunes only within fixed-size contiguous groups, which limits speedup~\cite{Hu2024a,Hu2024,castro2023venom}.

\textbf{Contributions.} To addresses these limitations, we introduce (Bl)ock (a)nd (S)parse (T)ransformers (\method{}). 
\method{} targets block sparsity in weight matrices, which we argue provides the best trade-off between hardware efficiency and model accuracy. Block sparsity offers sufficient structure for accelerator optimization while maintaining enough flexibility to preserve performance.
As shown in \Cref{fig:blast_overview}, the framework alternates between dense gradient updates and a pluggable block sparsification stage, executing sparse computations with a fused block-sparse matrix-matrix multiplication kernel. The design is agnostic to the pruning method. In this work, we instantiate two block pruning strategies, though other methods can be integrated without modification.
Our kernel outperforms vendor dense~\cite{NVIDIA_cuBLAS_2025}, and sparse~\cite{Abdelfattah2024,cusparse,hipsparse} libraries.
In our evaluations, \method{} reaches up to 95\% sparsity with negligible accuracy loss, achieves up to 521$\times$ kernel speedup over state-of-the-art SpMM, up to 23$\times$ over dense GEMM, 1.6$\times$ end-to-end speedups, and 2.2$\times$ distributed inference speedups on the Llama~3 family.

In summary, this paper makes the following contributions:
\begin{itemize}
    \item \textit{Framework:} We introduce \method{}, a block sparse neural network framework designed from the ground up for hardware efficiency. \method{} provides a unified, end-to-end solution for efficient machine learning, covering the full model lifecycle;
    \item  \textit{Kernel:} We develop a high-performance, scalable implementation of a custom block-sparse SpMM kernel optimized for modern GPU architectures;
    \item \textit{Evaluation:} We conduct an extensive empirical evaluation across 8 datasets and 11 model architectures, outperforming state-of-the-art kernels while preserving accuracy across tasks.
\end{itemize}

\section{Background}
\label{sec:background}
We now review the key components of the Transformer-based models. We also analyze them from the scope of sparsification strategies, and lay the foundations for \method{}.

\subsection{Transformer-based ML models}
\label{sec:transformer-based-ml-models}

The Transformer architecture~\cite{Vaswani2023} has emerged as the foundational building block of state-of-the-art large language models~\cite{Touvron2023a,Deepseekai2025,Geminiteam2024}, vision models~\cite{dosovitskiy2020image}, models for science~\cite{jumper2021highly,sonderby2020metnet}, and multimodal models~\cite{radford2021learning,ramesh2021zero}. Each block of the Transformer architecture comprises two main components: Attention and MLP, along with normalization layers (Figure \ref{fig:blast_overview}).
 
  \textbf{Attention.} The attention layer~\cite{shazeer2019fast, Vaswani2023, ainslie2023gqa, liu2024deepseek} captures relationships between tokens, where each token's output representation is computed based on the values of all other tokens in the sequence.

The \textbf{MLP} layer, on the other hand, multiplies the activation tensor $X$ directly with weight matrices. Depending on the architecture, typically there are two~\cite{radford2019language} or three~\cite{Touvron2023a} weight matrices. A Mixture-of-Experts (MoE)~\cite{shazeer2017outrageously} can be viewed as functionally analogous to a standard MLP layer, as both operate on each token \emph{independently}. In either architecture, the output state of a token is determined solely by its input state and the network's weights, without influence from other tokens in the sequence.  For the Llama3 family models, the MLP operator is defined as:
\begin{equation}
  Y = \text{MLP}(X) = \left(SiLU \left( X W_1 \right) \odot \left( XW_2 \right) \right) W_3
  \label{eq:mlp}
\end{equation}

where $SiLU$ represents the  Sigmoid Linear Unit activation function, $X$ represents the input to the MLP layer, and $W_1$, $W_2$, and $W_3$ are the weight matrices. The MLP layer accepts a tensor $X \in \mathbb{R}^{n \times s \times e}$ of size $n \times s \times e$, where $n, s, e$ are batch size, sequence length, and embedding size, respectively. Then the model performs a series of expansions and contractions on this input $X$ to generate the output $Y$ as shown in \Cref{eq:mlp}.

\subsection{Model Sparsification}
\label{sec:bac_sparse}

Related works on sparsification in deep learning suggest that no single method is universally applicable across all scenarios~\cite{hoefler2021sparsity}. Various techniques have been explored, from unstructured weight pruning~\cite{Frankle2019,prasanna2020bert,gordon2020compressing,Chen2020,guo1909reweighted,sanh2020movement}, which eliminates redundant individual connections, to structured approaches that remove entire neurons~\cite{polyak2015channel,hu1607network,pan2016dropneuron,lee2018snip}, filters~\cite{luo2020autopruner,luo2017thinet,he2019filter,you2019gate}, or methods that sparsify attention~\cite{michel2019sixteen,xu2025xattention,voita2019analyzing,mccarley2020structured,yun2020n,parmar2018image,guo2019star,child2019generating,beltagy2020longformer,zaheer2020big,li2020sac,zhao2019explicit}. 
Recently, XAttention~\cite{xu2025xattention} has proposed block sparse attention with antidiagonal scoring, and NSA~\cite{yuan2025native} introduces a natively trainable sparse attention mechanism, both for long-context Transformers.

However, empirical studies indicate that sparsifying MLP weights tends to be more robust, leading to substantial reductions in memory footprint and improvements in parallelization~\cite{shazeer2017outrageously,lepikhin2020gshard,fedus2022switch,huang2024toward}.  In contrast, approaches focused on sparsifying attention components mainly reduce the size of transient storage during computation without significantly lowering the overall parameter count. These observations underscore the need for flexible, multi-strategy sparsification approaches that balance efficiency gains with the preservation of network trainability and generalization.

In this work, we propose a \emph{general} solution that is applicable across the entire range of current and future Transformer architectures. 
Our proposed solution remains accurate as sparsity in the MLP layers increases, thereby ensuring scalable performance for the future of AI. Although our experiments focus on MLP sparsification, the framework accommodates sparsity in any linear layer.

\section{\method{} - (BL)ock (S)parse (T)ransformers}
\label{sec:blast}
\subsection{Overview}
We now describe \method{}, our novel framework for improving the pre-training and inference execution time, while also reducing memory footprint. The sparsification algorithms are described in \Cref{sec:blocked-prune-and-grow} and \Cref{sec:ollm}, and the BSpMM kernel in \Cref{sec:kernel_section}.

\subsection{Blocked Prune and Grow (\png{})}
\label{sec:blocked-prune-and-grow}

Magnitude pruning can expose early subnets during pretraining~\cite{Frankle2019,Chen2021a,Chen2021,You2022}. RigL improves accuracy by regrowing connections using gradients~\cite{Evci2021,Evci2022}. These methods yield unstructured sparsity that is hard to accelerate on GPUs due to bandwidth limits~\cite{Abdelfattah2024,naumov2010cusparse,hipsparse}. We adapt prune and grow to block-sparse weights in Transformer MLPs, as shown in \Cref{fig:blast_overview}, and refer to it as \emph{\png{}}.

The \png{} algorithm gradually reduces the number of parameters of the weight matrix $W$ until a sparsity $s_{max}$ has been achieved over $m$ iterations. Listing~\ref{lst:loop-prune-and-grow} shows the algorithm with a regular training loop as Python code. The training loop and optimizer are executed in \texttt{forward\_and\_backward\_step()}. The masks are updated once every $step\_size$ iterations. The masks for each weight matrix are stored in boolean tensors and applied to the weight matrices in the \texttt{prune\_weights()} step.

\begin{lstlisting}[language=Python, label=lst:loop-prune-and-grow, caption=Training loop of the blocked prune and grow algorithm. The actual grow and prune procedure is implemented in the function \texttt{generate\_masks()}.]
for iteration in range(train_iters):
  forward_and_backward_step()
  if iteration % step_size == 0:
    masks = generate_masks()
  prune_weights(masks)
\end{lstlisting}

\Cref{fig:blast_overview} shows one iteration of \texttt{generate\_masks()} on weight $W_i$ and gradient $G_i$. We partition $W_i$ and $G_i$ into $b\times b$ blocks and apply $S(\cdot)$. Each block is scored by its Frobenius norm. We drop the lowest scoring blocks until we reach sparsity $s_i$, as given in \Cref{eq:cubic-sparsity}. Zeroed blocks are shown in white in the figure. Next, we take the set difference of the nonzero block indices between $S(W_i)$ and $S(G_i)$. The resulting indices, shown in red in matrix $D_i$, are regrown into $S(W_i)$, which yields the updated weight $W_{\text{new}}$.
 Newly regrown blocks are initialized to zero so that they do not
affect the current transform and are updated during subsequent optimization. In practice, we keep boolean masks for pruning and growth and form $W_{\text{new}}$ in \texttt{prune\_weights()}.

The weight sparsity $s_i \in [0, 1)$ at iteration $i$ is determined by \Cref{eq:cubic-sparsity}~\cite{Zhu2017}. 
 \begin{equation}
  s_i = s(i) = s_{max} + (s_{init} - s_{max})  \left(1 - \frac{i}{m - d}\right)^3 ,
  \label{eq:cubic-sparsity}
\end{equation}
 where $s_{init}$ represents the initial sparsity, which determines the sparsity at the beginning of the training, $m$ is the maximum number of training iterations, and $d$ is a decay term that controls how fast the maximum sparsity $s_{max}$ is reached. Cubic scaling provides for a smooth and gradual increase in sparsity.

Unlike pruning algorithms that use straight-through estimators~\cite{Hu2024a,Hubara2021,Bengio2013}, our method prunes the weight matrices directly. Consequently, the same pruned weight matrix is applied during both forward and backward propagation. During training, the dense weight and gradient matrices are kept intact, while the pruned weight $W_{new}$ is stored in the blocked compressed sparse column (BCSC) format~\cite{im2000optimizing} 
for forward and backward propagation. Although this leads to higher memory consumption during training (which gradually decreases as the sparsity $s_i$ increases), it ultimately yields improved speedup and reduced inference memory usage, as shown in \Cref{sec:results}.

\subsection{oLLM}
\label{sec:ollm}

We propose a method called ``oLLM'' that uses second-order gradient information for pruning weight blocks within layers, inspired by the optimal brain surgeon method \cite{Hassibi1993}. The OBS based sparsification process consists of four steps: first, estimate the model's curvature; second, assign sensitivity scores to each weight based on this curvature; third, prune weights with the lowest sensitivity; and fourth, update the remaining nonzero weights to preserve the curvature estimated by the second-order gradient.

The core principle of OBS is to estimate the sensitivity $\mathcal{L}_k$ of the $k$-th weight using the diagonal element $H^{-1}_{kk}$ of the inverse Hessian matrix $H^{-1}$, as shown in \Cref{eq:obs-sensitivity}. Weights with lower sensitivity can be pruned with minimal accuracy loss.
\begin{equation}
    \mathcal{L}_k = \frac{1}{2}\frac{W_k^2}{[H^{-1}]_{kk}}
    \label{eq:obs-sensitivity}
\end{equation}

The remaining weights are then updated as shown in \Cref{eq:obs-weight-update}, where $e_k$ is a basis matrix for selecting the corresponding column of $H^{-1}$. The size of the Hessian matrix is a square of the number of weights, which makes it intractable to store and invert in a finite amount of time. To overcome this limitation, the diagonal inverse Fisher approximation\cite{Gupta2018,Kurtic2022,Grosse2016,Vanderouderaa2024} can be used for approximating the inverse Hessian $H^{-1}$ to save space and computation time.

\begin{equation}
    \Delta W = - \frac{W_k}{[H^{-1}]_{kk}}H^{-1}e_k
    \label{eq:obs-weight-update}
\end{equation}

\begin{figure}
    \centering
    \includegraphics[width=\linewidth]{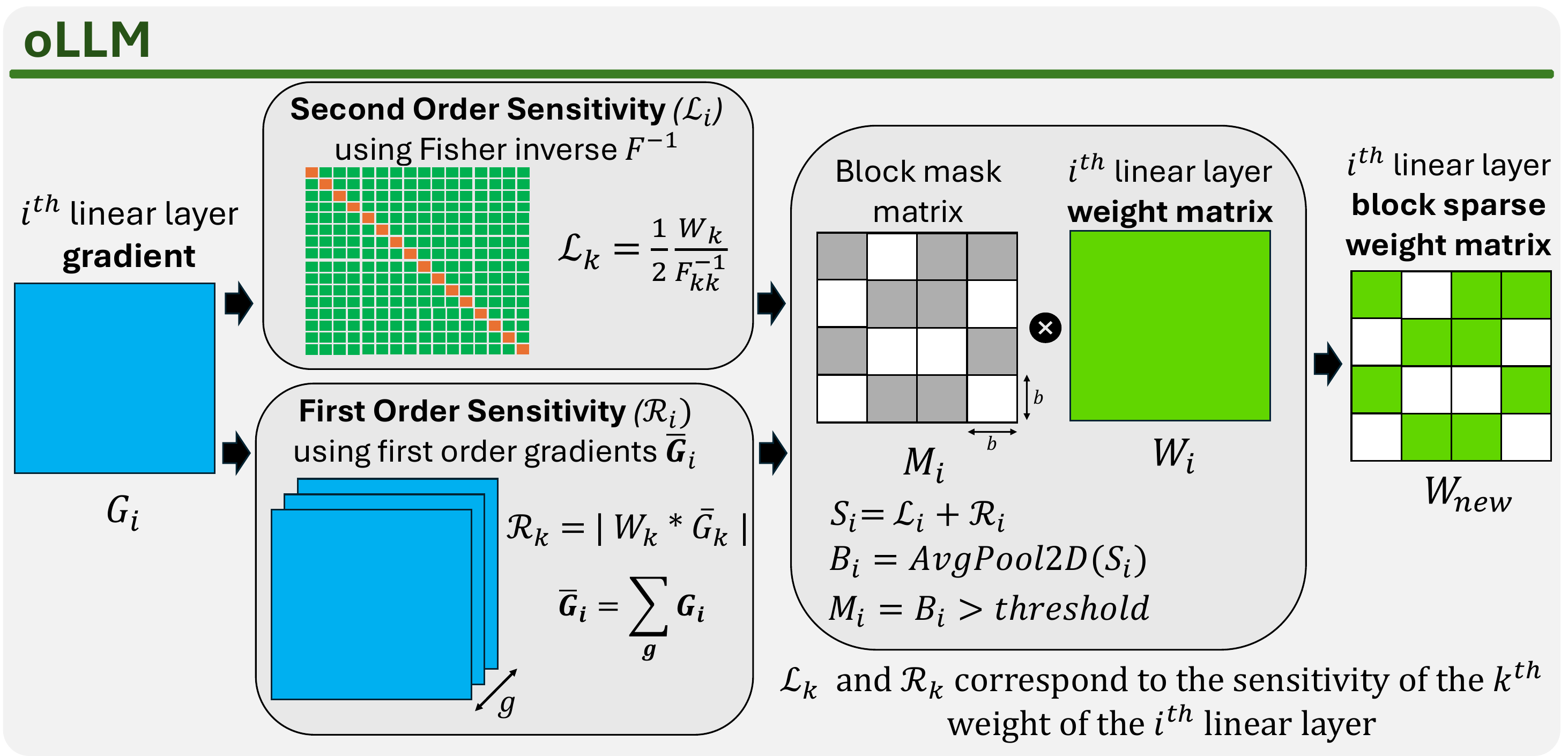}
    \caption{The blocked oBERT accumulates weight gradients $G_i$ over many forward and backward iterations to generate the sensitivity scores $S_i$ using the Fisher inverse matrix. $S_i$ is then used for generating a blocked mask $M_i$. The blocked weight matrix $W_{new}$ is generated with an element-wise multiplication of the weight matrix $W_i$ with $M_i$.}
    \label{fig:blocked-obert}
\end{figure}
 
Second order pruning methods such as oBERT~\cite{Kurtic2022} assume that the first order gradient is already at the minimum. On the contrary, methods such as LPViT~\cite{xu2024lpvit} incorporate the first order gradient information into the sensitivity score calculation and show better accuracy. \Cref{eq:first-order-sensitivity} shows the calculation of the sensitivity $\mathcal{R}_k$ using the first order gradient for the $k^{th}$ weight in weight matrix $W$. $\bar{G}$ is the accumulated first order gradient over $g$ iterations as shown in \Cref{fig:blocked-obert}. \Cref{eq:hybrid-sensitivity} then computes the total sensitivity $S_k$ using $\mathcal{L}_k$ and $\mathcal{R}_{k}$.

\begin{equation} 
\mathcal{R}_{k} = |W_k \cdot \bar{G}_k| 
\label{eq:first-order-sensitivity} 
\end{equation} 

\begin{equation} 
\mathcal{S}_{k} = \mathcal{L}_{k} + \mathcal{R}_{k} 
\label{eq:hybrid-sensitivity} 
\end{equation}

The weight sensitivity $S_k$ of each weight is combined to form the sensitivity matrix $S_i$ corresponding to each weight in the weight matrix $W_i$ as shown in \Cref{fig:blocked-obert}. $S_i$ is then split into blocks of size $b \times b$, and the sensitivity score of the block is obtained with 2D average pooling~\cite{zafar2022comparison}. The weights corresponding to blocks with the least block sensitivity are then pruned to obtain the block sparse weight matrix $W_{new}$.

\subsection{BSpMM Kernel}
\label{sec:kernel_section}
We now introduce our custom BSpMM kernel, the elementary algebraic building block that secures maximal speedup over dense baselines.
The BSpMM kernel, while being part of our \method{} framework, can also be used independently as a stand-alone SpMM kernel, paving 
the road for fast, sparse linear algebra kernels across various domains.
We then further extend it to the MLP blocks inside Transformer architectures to capture the most common compute pattern in ML workloads.

\subsubsection{Kernel Outline}

\begin{figure}
    \centering
    \includegraphics[width=0.9\linewidth]{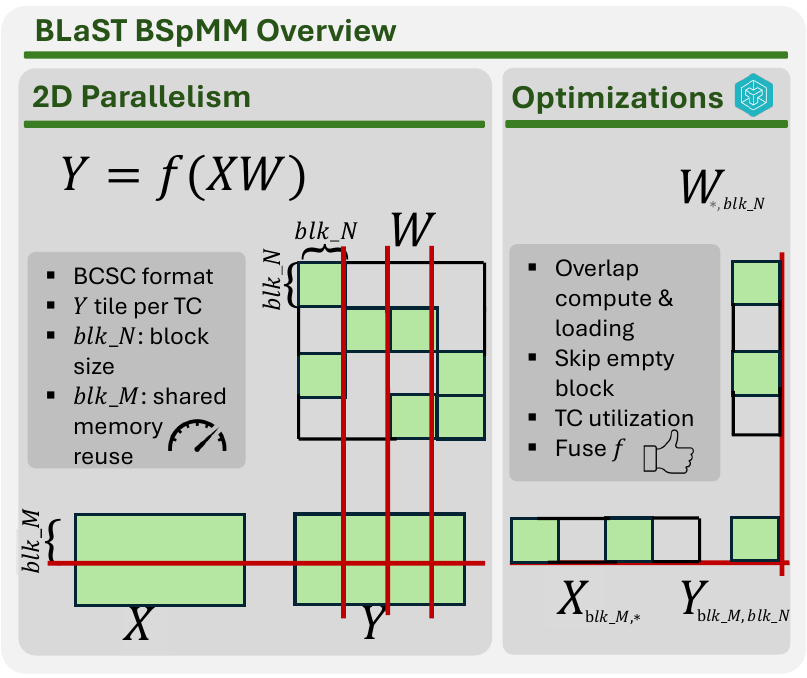}
    \caption{Overview of \method{} BSpMM. The kernel uses a blocked Compressed Sparse Column (BCSC) layout derived from sparsified neural network weights. \method{} applies block-level, bottom-up 2D parallelism in Triton to maximize GPU utilization.}
    \label{fig:kernel_overview}
\end{figure}

Different internal storage formats are preferred depending on whether the sparse matrix $W$ is multiplied from the left ($Y = XW$) or from the right ($Y = WX$). Since the PyTorch implementation of a \texttt{torch.nn.Linear} linear layer assumes multiplying from the left, for brevity, we explain only this variant. The $Y=WX$ kernel is symmetrical, so we omit it from this discussion.

\Cref{fig:kernel_overview} illustrates the high-level design of our \method{} BSpMM kernel. We build on prior work such as SMaT~\cite{okanovic2024high}, then focus on a block-sparse path tailored to our setting.

We store sparsified weights in a BCSC format. The launch grid is derived from BCSC so that we create work only for output tiles that have contributing nonzero blocks in $W$. Each output tile in $Y$ is assigned to one warp and driven by a single Tensor Core. We refer to this as bottom-up 2D parallelism because tiles exist only where $W$ has nonzero blocks. This reduces load imbalance and avoids work on empty regions.

For each output tile $Y_{i,j}$ our kernel calls low-level CUDA MMA (matrix multiply accumulate) instructions on the warp’s Tensor Core multiplying nonzero blocks $W_{:,j}$ with corresponding dense blocks $X_{i,:}$. The block size matches the MMA shape, enabling direct register-level fragments. We stream blocks with asynchronous copies into shared memory, and double buffer to overlap data movement with compute. Because iteration follows the BCSC column pointers, the kernel touches only existing blocks and never computes on absent tiles, which eliminates fill-in and keeps the Tensor Core pipelines busy.

The carefully tuned rectangular block shape results from optimizing shared memory and register usage, while minimizing accuracy loss due to overly aggressive sparsification. The two block dimension, \textit{blk\_M},  and \textit{blk\_N}, are optimized for compute performance, and model accuracy, respectively. Note that \textit{blk\_N} corresponds to block parameter $b$ introduced in \Cref{sec:blocked-prune-and-grow}:

\begin{enumerate}
    \item \textit{blk\_N} determines the sparse block size. When the block is smaller, the sparsification strategy has more freedom to choose the best blocks to prune, resulting in a more fine-grained masking process. Conversely, larger blocks enable more efficient computation. We empirically select the largest block sizes that maintain accuracy, with a median relative accuracy drop of 2.24\%. 
    \item \textit{blk\_M} controls the number of rows from the dense matrix that participate in the thread-block level reduction, thus increasing the reuse of data loaded into shared memory from the sparse block. We use the COSMA analytical compute model~\cite{kwasniewski2019red} to derive the optimal block size, given hardware constraints such as the number of available registers, shared memory capacity, and warps per threadblock.
\end{enumerate}

\subsubsection{Implementation}
To secure architecture-agnostic implementation and kernel portability, BSpMM is implemented in Triton~\cite{tillet2019triton}. Triton is 
well-suited for block computations, using advanced heuristics to automatically generate efficient accelerated code for both NVIDIA and AMD GPUs.
It seamlessly utilizes all modern features of the target hardware, such as asynchronous pipeline loads, Tensor Memory Accelerators (TMAs)~\cite{Nvidia2022}, or
Tensor Core~\cite{Markidis2018} computations. Furthermore, Triton rearranges threads within a thread block (called \emph{program instance}) to avoid bank conflicts,
vectorize memory accesses, and align registers for Tensor Core's fragments.

The main challenge to implement a high-performance BCSC-based, batched SpMM kernel is to handle inherent data indirection efficiently.
Compared to dense $Y=XW$, all memory accesses are known at the compile time, and we can create a provably optimal static schedule 
with compile time tilings, data distribution, and memory pipelining~\cite{ziogas2022deinsum}. The sparse data structure prevents the static
access pattern of the dense matrix $X$, since, to achieve data movement optimality, only necessary blocks of $X$ can be loaded, and the sparsity structure of $W$ determines this. We implement this in Triton by determining the dynamic range of accesses to $X$ with pointer algebra on the  
\texttt{blk\_col\_ptr} pointer array. The main part of the kernel is presented in Listing~\ref{lst:spmm-kernel}. The Triton API can be found in its documentation~\cite{tillet2019triton}.

\begin{lstlisting}[language=Python, label=lst:spmm-kernel, caption={Main compute loop of the BSpMM kernel}, float, floatplacement=tb]
for k in range(num_blocks):
    # load BCSC block 
    W_block = tl.load(block_vals_ptrs + k * blk_stride)
    # calculate the block row of the current BCSC block
    row_idx = tl.load(blk_row_idx + k)
    # move X_ptrs to the correct block
    X_ptrs = X_ptrs_base + row_idx * blk_K * stride_xk
    # load dense batched matrix (activation tensor X)
    X_block = tl.load(X_ptrs)
    accumulator = tl.dot(X_block, W_block, accumulator) 
    # Advance the ptrs to the next K block.
    blk_vals += blk_M * blk_stride
\end{lstlisting}

\subsubsection{Block Sparse MLP}
A Transformer MLP uses two affine projections with a element-wise nonlinearity in between, such as ReLU, GELU, or SiLU. 
While tensor contractions are usually compute-bound (even in the sparse case), nonlinear operators are memory-bound.
The arithmetic intensity, the ratio of arithmetic operations per loaded element, is a small constant, usually 1 or 2.
Therefore, to prevent a memory bottleneck, nonlinear operators have to be fused with the compute-bound operations.

Our optimized BSpMM kernel for MLPs fuses the nonlinearities that follow the SpMM operator into one kernel. Our implementation supports
all common nonlinearities supported by Triton, further improving performance and reducing data movement.

\section{Evaluation}
\label{sec:evaluation}
We conduct a comprehensive set of experiments to evaluate the performance, and accuracy of \method{}.
Our experiments span 8 different datasets in both text and vision domain, across 11 different Transformer architectures.

In \Cref{sec:results}, we train the models on a diverse set of benchmark datasets, including OpenWebText~\cite{Gokaslan2019OpenWeb}, CIFAR-10~\cite{krizhevsky2009learning}, DialogSum~\cite{paperno2016lambadadatasetwordprediction}, and five tasks from the GLUE benchmark~\cite{wang2019gluemultitaskbenchmarkanalysis}.

Furthermore, we conduct experiments across both encoder and decoder-only Transformer architectures, along with their respective parameter counts: Llama3.2 \cite{Panferov2025} (1.59B), Qwen3~\cite{Yang2025a} (1.7B), BERT~\cite{Devlin2019} (110M), GPT2-small \cite{Radford2019} (124M), GPT2-medium \cite{Radford2019} (355M), GPT2-large  \cite{Radford2019} (774M), GPT2-XL  \cite{Radford2019} (1.44B), ViT-B/16 \cite{dosovitskiy2020image} (86M), ViT-B/32  \cite{dosovitskiy2020image} (86M), ViT-L/16  \cite{dosovitskiy2020image} (307M), ViT-L/32  \cite{dosovitskiy2020image} (307M). In all our experiments, we adhere to the scientific benchmarking guidelines proposed by Hoefler et al.~\cite{hoefler2015scientific}. That is, when reporting speedups we report the single parallel process as the base case, and report the absolute performances of base cases in the appendix.

\paragraph{\textbf{Hardware Infrastructure}}
We evaluate \method{} on a compute partition, where each node is equipped with four NVIDIA Grace Hopper GH200 superchips interconnected with HPE Cray Slinghot-11. Unless otherwise stated, measurements use a single GH200 superchip. Multi-node configurations, when reported, scale up to 4 nodes (16 GH200s).

\paragraph{\textbf{Software Stack}}
All experiments were executed using Pytorch v2.7.0, CUDA v12.8, Megatron-core v0.10.0, Apex v0.1, Triton v3.0.0, and Transformers\_engine v2.0.0.\\

\noindent
Code and documentation are made available in the code repository: \url{https://github.com/spcl/blast}.

\section{Results}
\label{sec:results}
\subsection{\method{} BSpMM Kernel Comparison}
\label{sec:bspmm-kernel-comparison}
\begin{figure*}
    \centering
    \includegraphics[width=\linewidth]{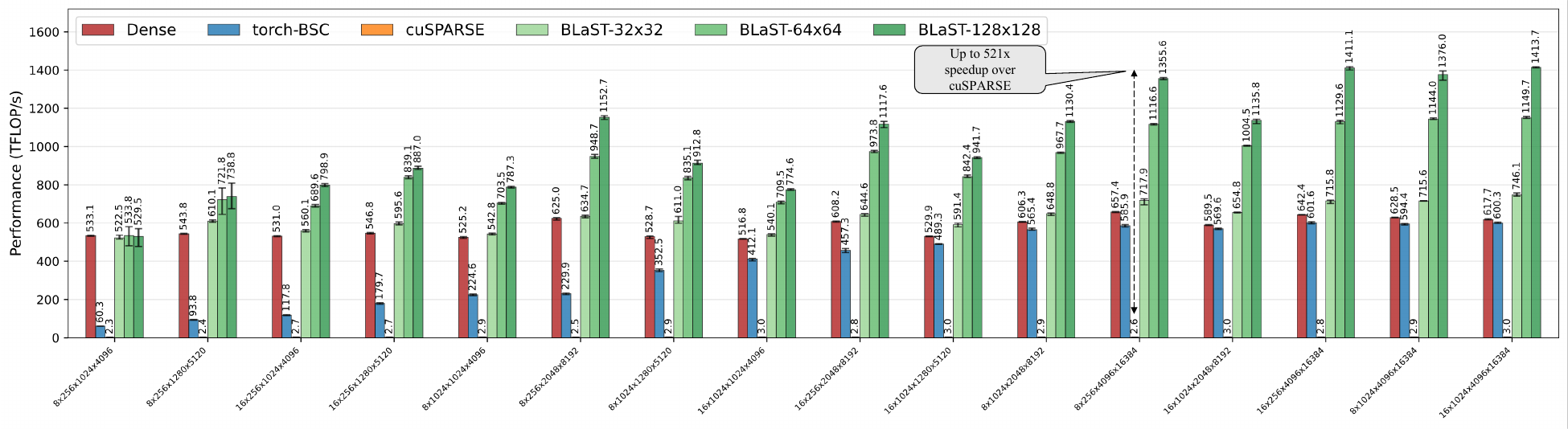}
    \caption{Throughput of \method{} BSpMM compared to state-of-the-art kernels for various combinations of batch size, embedding and sequence length typically found in the networks that we benchmark. 
    \method{} BSpMM can outperform all vendor implementations by a wide margin for all block sizes.}
    \label{fig:kernel_comparison}
\end{figure*}
In this section, we benchmark the performance of our BLaST BSpMM kernel against state-of-the-art matrix multiplication kernels. We compare against PyTorch's dense matrix multiplication implementation~\cite{paszke2019pytorch}, which utilizes CUTLASS and cuBLAS libraries, PyTorch with Block Sparse Connectivity (BSC) sparsity format, as well as cuSPARSE~\cite{naumov2010cusparse} for 
sparse matrix operations. \Cref{fig:kernel_comparison} shows the throughput of \method{} (fixed 70\%) for various matrix shapes over different matrix shapes, varying the embedding, sequence, and hidden dimensions as described in the previous chapter. This shows that BLaST BSpMM consistently outperforms all state-of-the-art kernels across the evaluated configurations. In particular, BLaST is capable of outperforming cuSPARSE by up to 521x, showcasing the significant computational advantages of our kernel.

\subsection{\method{} BSpMM Performance}
\label{sec:mlp-layers-speed}

The previous section demonstrated the performance of our BSpMM kernel against other block SpMM kernels. Results show that dense kernels from cuBLAS and CUTLASS consistently rank second to our implementation. We now analyze the speedup achieved by the \method{} BSpMM kernel relative to state-of-the-art dense matrix multiplication algorithms, reporting speedup as the ratio
\[
\frac{\min\left( t_{CUTLASS},\, t_{cuBLAS} \right)}{t_{\method{}}},
\]
where \(t_{CUTLASS}\) represents the execution time using the CUTLASS library, \(t_{cuBLAS}\) denotes the time for cuBLAS, and \(t_{\method{}}\) is the execution time for our method. In all experiments, we present the results for \textit{blk\_M = blk\_N} \(\in \{16, 32, 64, 128\}\).

\begin{figure}
    \centering
    \includegraphics[width=\linewidth]{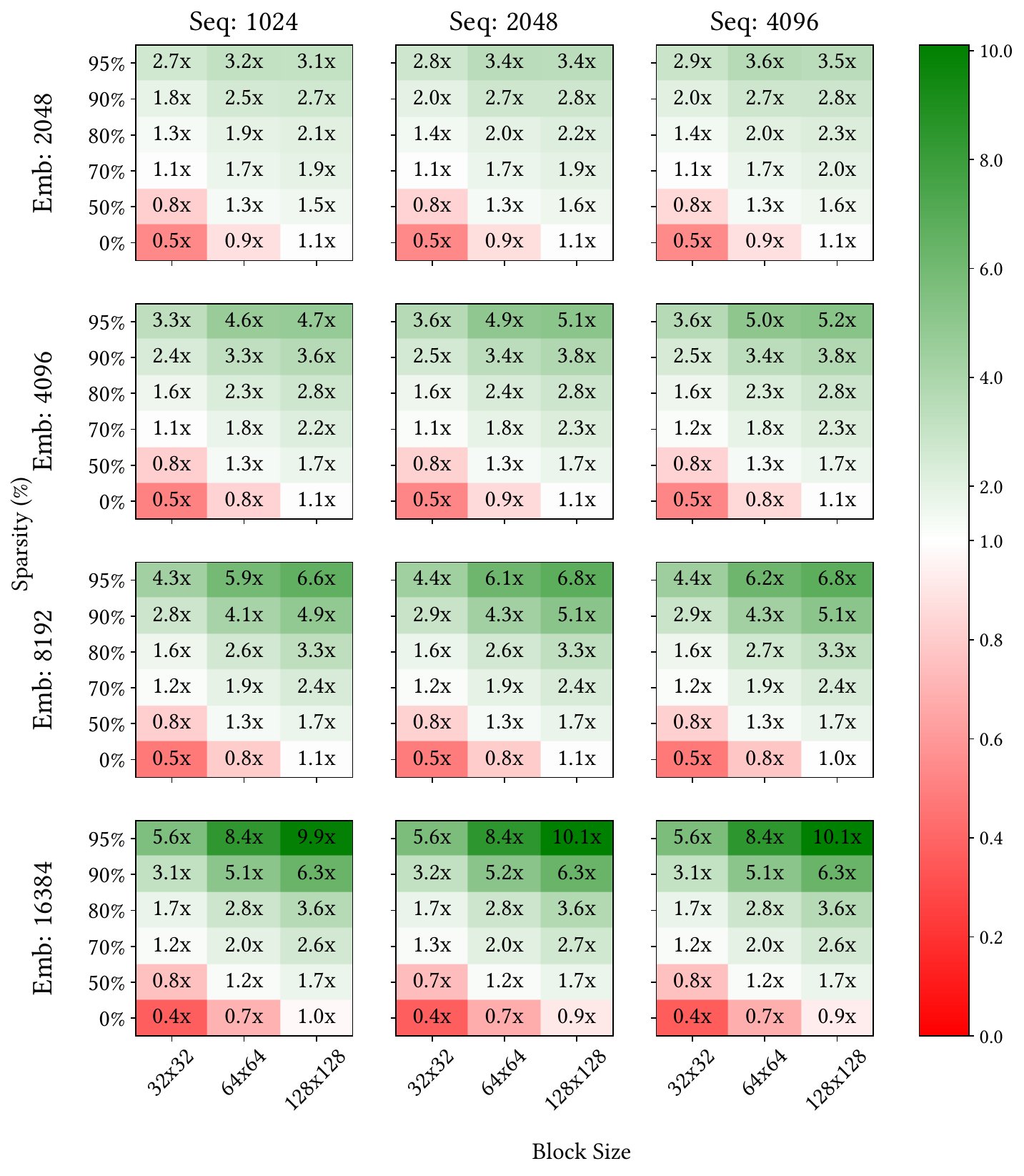}
    \caption{Ablation study on the BSpMM kernel speedup for various block sizes and embedding sizes, showing scaling up results for LLMs.}
    \label{fig:kernel_speedup}
\end{figure}

\Cref{fig:kernel_speedup} shows the speedup of the BSpMM kernel when multiplying individual matrices corresponding to typical embedding and context length sizes in LLMs. In a matrix multiplication routine characterized by the dimensions \(M\), \(N\), and \(K\), the \(Emb\) dimension corresponds to \(K\), \(Seq\) corresponds to \(M\), and \(N\) is set to \(4 \times Emb\). This indicates that as both the \(Emb\) and \(Seq\) dimensions increase, the speedup also increases. \method{} BSpMM is faster by as much as 16.7x for \(Emb = 32,768\), \(Seq = 1,024\), and sparsity 95\%.  We observe that, in general, higher performance is achieved with a larger block size. Moreover, we observe that \method{} SpMM with a 128×128 block size can even outperform cuBLAS by 1.1x in the fully dense case (0\% sparsity). Note that \method{} BSpMM supports all data types supported by Triton~\cite{tillet2019triton}. Due to the space constraints, we include the results only for the most popular data type in ML workloads: BF16. Using lower precision, e.g.
\texttt{torch.float8\_e5m2}, we measured up to 23x speedup for $Emb$ = 32,768, $Seq$ = 1,024, and sparsity 95\%.

\begin{figure}[ht]
    \centering
    \includegraphics[width=\linewidth]{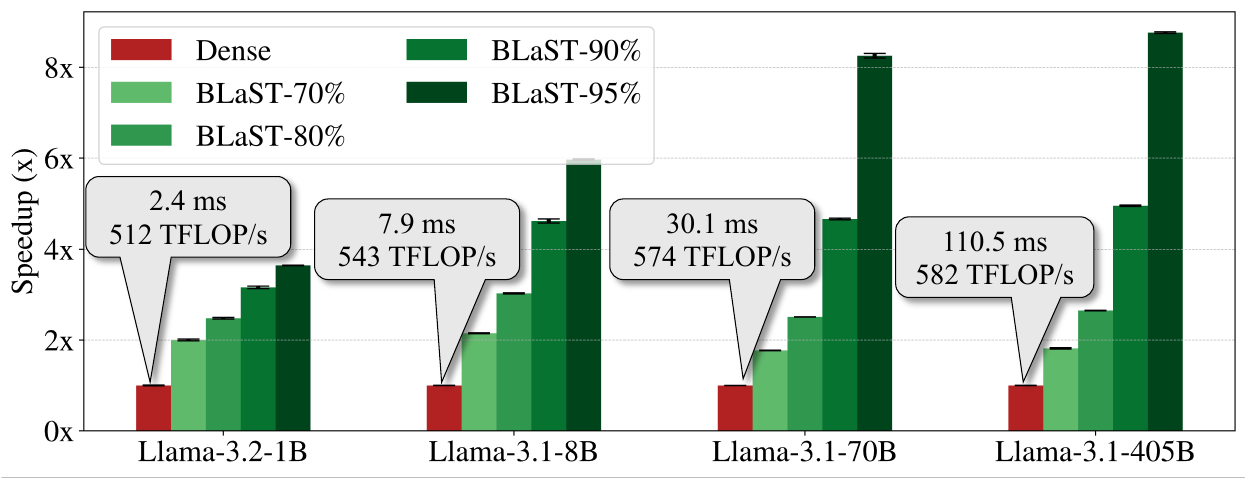}
    \caption{MLP layer speedups for the Llama family with BLaST 128x128, reaching 8.8× for Llama-3.1 405B}
    \label{fig:llama_fd}
\end{figure}

\begin{figure}[ht]
    \centering
    \includegraphics[width=\linewidth]{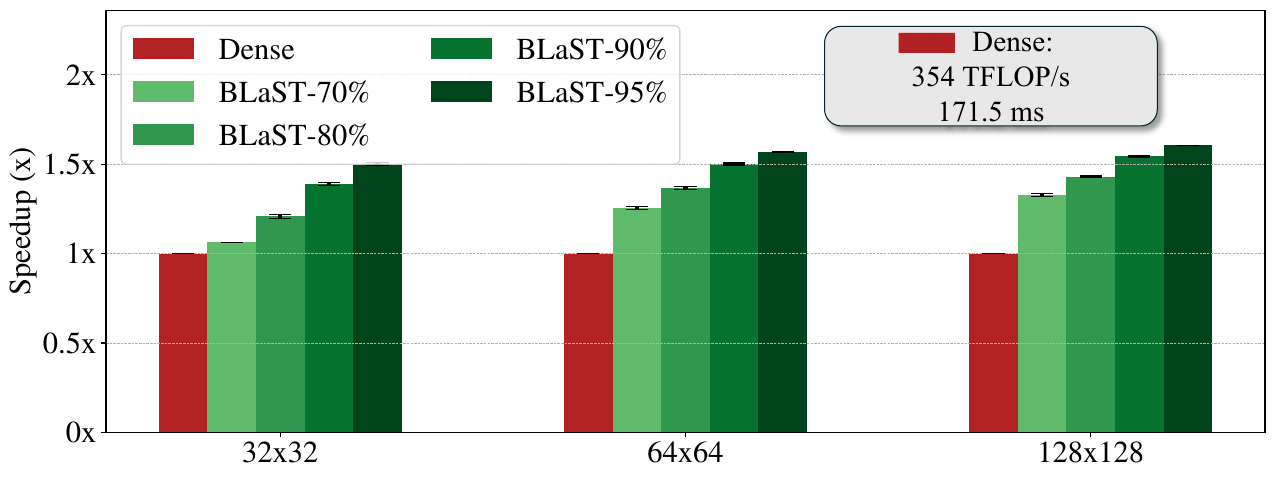}
    \caption{End-to-end inference speedup for Llama-3.2 1B for various block sizes and sparsity levels.
For 70\% sparsity, BLaST yields 1.3x speedup, and for 95\% sparsity, BLaST reaches up to 1.6x speedup.}
    \label{fig:llama_inference}
\end{figure}

Lastly, we evaluate the speedup of \method{} BSpMM in the Llama model family, ranging from the Llama-3.2 1B to the Llama-3.1 405B. \Cref{fig:llama_fd} demonstrates the speedup in the MLP blocks. We observe that even 70\% sparsity in the MLP blocks is sufficient to achieve a 2.15x speedup. As the models scale up to 405B parameters, we achieve up to an 8.77x speedup for 95\% MLP sparsity. In \Cref{fig:llama_inference}, we show the end-to-end speedup in the inference of the Llama-3.2 1B model across different block sizes. One can achieve up to a 1.6x speedup even for a smaller model such as the Llama-3.2 1B using our proposed BSpMM. In conclusion, these experiments demonstrate that \method{} represents a promising step towards scaling models to the trillion-parameter era.

\subsection{Distributed Results}
\begin{figure}[ht]
    \centering
    \includegraphics[width=\linewidth]{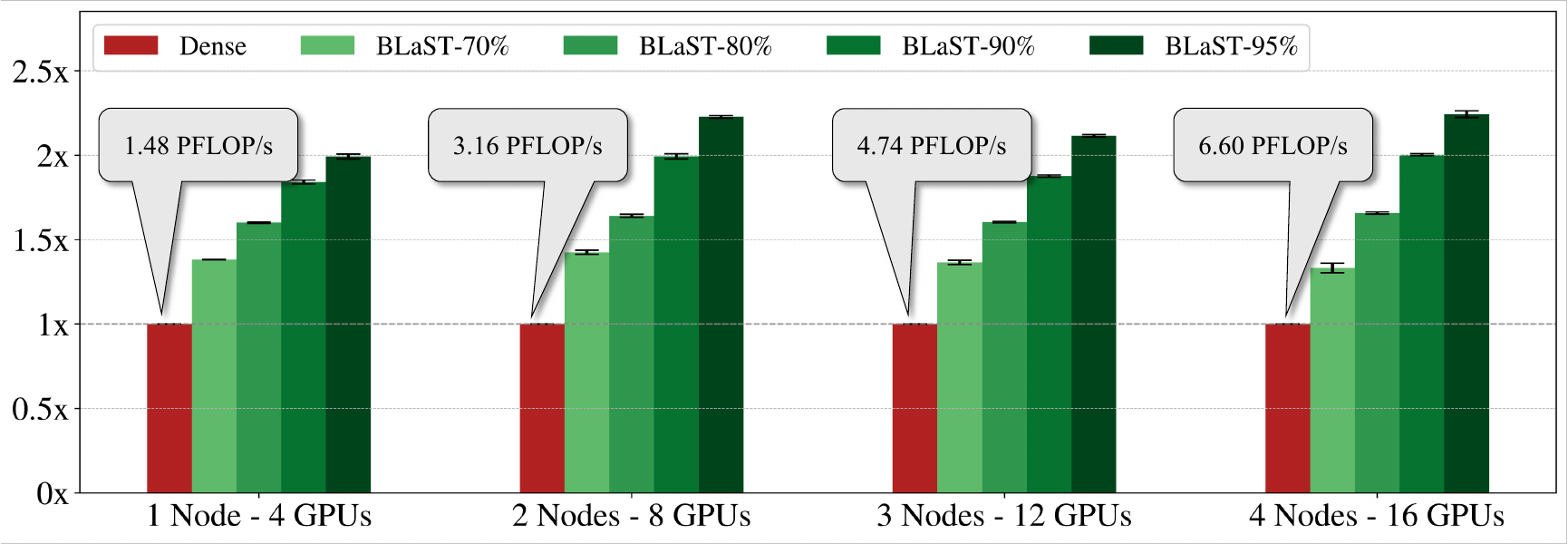}
    \caption{Speedup of \method{} in the distributed setting using Megatron-LM. We scale the Llama 3 architecture up to 16 NVIDIA GH200 GPUs. As both model size and computational resources scale up, \method{} consistently outperforms the dense baseline. \method{}-95\% achieves a speedup of 2.2x on 16 GPUs.}
    \label{fig:distributed_results}
\end{figure}

In this section, we demonstrate the results of our method in a distributed setting. 
We benchmark the results on up to four nodes, where each node has four NVIDIA Grace Hopper GH200 superchips interconnected with HPE Cray Slinghot-11.
\method{} can be adapted to all versions of parallelism used in deep learning~\cite{ben2019demystifying}, since it is fundamentally a matrix multiplication method. For data parallelism, \method{} scales linearly with the number of GPUs as the model is replicated across each device, maintaining the same computational efficiency per GPU. The method naturally extends to tensor parallelism, where matrix operations are partitioned across devices, and model pipeline parallelism, where different layers are distributed across GPUs while preserving the sparsity patterns within each pipeline stage. Our method can be seamlessly integrated with any framework for distributed training. We evaluate our results using one of the most popular frameworks, Megatron-LM~\cite{shoeybi2019megatron}, which provides efficient implementations of transformer models with support for various parallelization strategies including tensor, pipeline, and data parallelism. \Cref{fig:distributed_results} demonstrates our improvement over the dense baseline. We scaled up the Llama3 architecture according to the scaling laws~\cite{Kaplan2020}, reaching 540B parameters for 16 GH200 NVIDIA GPUs. \method{} is 2.2x end-to-end faster compared to the dense model for 16 GPUs spread across 4 nodes for 95\% sparsity in the MLP layers.
Our results show that \method{} consistently outperforms the dense baseline across all distributed configurations.

\subsection{\method{} for Inference}
\label{sec:fine-tuning-inference}

Pretraining neural networks from scratch demands substantial computational resources. The widespread availability of pretrained LLMs
can be leveraged rather than retrain a network from scratch. This section reports model accuracy across tasks under block sparsity at inference. We organize results by task and evaluate multiple models, datasets, and block sparsification algorithms. The goal is to show that \method{} is effective across a broad range of practical LLM applications.

\subsubsection{Text Summarization}
\label{sec:summarization-task}

The DialogSum~\cite{Chen2021c} dataset tests whether an LLM can summarize a dialogue between two speakers. Table~\ref{tab:text-summary-dialogsum} shows the ROUGE-SUM\cite{Lin2004} score when using Llama3.2-1B~\cite{Touvron2023a}. and Qwen3-1.7B~\cite{Yang2025a} on DialogSum. Block size $1$ denotes fine-grained sparsity followed by block sparsity with block size ranging from 16 to 128.

The pruning algorithm influences the ROUGE-SUM scores achieved by the model for a given block size and target sparsity. oLLM, described in \Cref{sec:ollm}, improves scores for Llama3.2 and Qwen3.
For the blocked prune and grow algorithm (\Cref{sec:blocked-prune-and-grow}), fine-grained sparsity (i.e. block size 1) shows the worst ROUGE-SUM scores. Increasing the block size from 16 to 128 improves the score. For the oLLM algorithm, the best ROUGE-SUM score is obtained for block size 1, whereas a larger block size shows deterioration in the ROUGE-SUM score.

We found that a procedure of dense finetuning followed by sparsification repeated multiple times works best for this dataset. This cycle of dense-sparse fine-tuning is done twice for all data points, leading to a total of 4 epochs of fine-tuning over the dataset. The dense baseline is computed using a single epoch. While accurate sparsification requires four times the fine-tuning cost of the dense model, the substantial inference speedups demonstrated in \Cref{sec:fine-tuning-inference} justify this additional one-time investment, as inference constitutes the majority of LLM operational costs.
\begin{table}
\centering
\resizebox{\columnwidth}{!}{%
    \begin{tblr}{
      cells = {c},
      cell{1}{1} = {r=2}{},
      cell{1}{2} = {r=2}{},
      cell{1}{3} = {r=2}{},
      cell{1}{4} = {c=5}{},
      cell{1}{9} = {r=2}{},
      cell{3}{1} = {r=8}{},
      cell{3}{2} = {r=4}{},
      cell{3}{9} = {r=8}{},
      cell{7}{2} = {r=4}{},
      cell{11}{1} = {r=8}{},
      cell{11}{2} = {r=4}{},
      cell{11}{9} = {r=8}{},
      cell{15}{2} = {r=4}{},
      hline{1,3,11,19} = {-}{},
      hline{7,15} = {2-8}{},
    }
    \textbf{Model}  & \textbf{Pruning alg.} & \textbf{$s_{max}$} & \textbf{Block size ($b$)} &             &             &             &              & \textbf{Dense}      \\
                    &               &                         & \textbf{1}     & \textbf{16} & \textbf{32} & \textbf{64} & \textbf{128} &                \\
    {Llama3.2~\\1B} & \png{}(\ref{sec:blocked-prune-and-grow})  & 70\%         & 31.99       & 37.81    & 38.14       & 38.35    & 38.11        & 37.43      \\
                    &               & 80\%                    & 31.55          & 36.28       & 37.47       & 37.41       & 37.15        &                \\
                    &               & 90\%                    & 31.37          & 35.39       & 35.78       & 35.59       & 34.91        &                \\
                    &               & 95\%                    & 31.15          & 34.5        & 34.54       & 33.84       & 33.52        &                \\
                    & oLLM (\ref{sec:ollm})  & 70\%           & 38.64         & 37.33       & 37.15       & 36.9        & 37.13        &                \\
                    &               & 80\%                    & 37.83         & 36.58       & 36.43       & 36.4        & 36.7         &                \\
                    &               & 90\%                    & 36.93         & 34.93       & 35.16       & 33.57       & 33.45        &                \\
                    &               & 95\%                    & 35.89         & 33.15       & 33.5        & 31.85       & 31.68        &                \\
    {Qwen3\\1.7B}   & \png{}(\ref{sec:blocked-prune-and-grow})        & 70\%    & 30.73     & 38.68    & 38.89       & 38.35       & 38.28      &  41.22              \\
                    &               & 80\%        & 29.87          & 37.73       & 37.86       & 37.41       & 37.15        &                \\
                    &               & 90\%        & 30.53          & 36.48       & 35.78       & 35.59       & 34.49        &                \\
                    &               & 95\%        & 30.85          & 35.16       & 34.54       & 33.84       & 33.82        &                \\
                    & oLLM (\ref{sec:ollm})       & 70\%                    & 40.82          & 38.95       & 39.05       & 38.97       & 38.85        &                \\
                    &               & 80\%      & 40.11          & 38.34       & 37.69       & 37.57       & 36.97        &                \\
                    &               & 90\%      & 38.75          & 36.68       & 35.93       & 35.73       & 35.92        &                \\
                    &               & 95\%      & 37.52          & 35.34       & 34.52       & 34.75       & 34.52        &                
    \end{tblr}
}
\caption{ROUGE-SUM~\cite{Lin2004} score for text summarization for the DialogSum dataset using Llama3.2 and Qwen3. ROUGE-SUM measures the longest common subsequence (LCS) between a generated summary and reference summary, evaluating how well the generated summary captures the content and ordering of the reference.}
\label{tab:text-summary-dialogsum}
\end{table}

This results can be attributed to the way in which these algorithms perform sparsification. The P\&G is essentially a magnitude pruning algorithm that first prunes the weights with the least magnitude, and then grows back some of them using the heuristic of the weight gradient with large magnitudes. This heuristic provides only a rough approximation of the model's curvature. Therefore, increasing the block size has the impact of incorporating weights that otherwise might have been removed when using fine-grained sparsity. On the other hand, oLLM uses second order curvature information to estimate weight importance more accurately. As a result, it can identify important weights with fine grained sparsity. Using 2D average pooling to aggregate sensitivity scores introduces inaccuracy by assuming that entire weight blocks can be pruned based on their average sensitivity. This approximation reduces the ROUGE-SUM score for oLLM under blocked sparsity.

\subsubsection{Text Classification (GLUE)}
\label{sec:text-classfication-glue}

The GLUE benchmark~\cite{wang2019gluemultitaskbenchmarkanalysis} assesses the general language understanding capabilities of models. Specifically, we evaluate on five task: Recognizing Textual Entailment (RTE), the Multi-Genre Natural Language Inference Corpus (MNLI), The Stanford Sentiment Treebank (SST-2), the Winograd Schema Challenge (WNLI), and report the average score calculated as the arithmetic mean of all other task specific metrics. 

In \Cref{tab:glue-benchmark-llama32} we compare the performance of the Llama 3.2-1B model when sparsified with the P\&G algorithm (\Cref{sec:blocked-prune-and-grow}) under different levels of sparsity, and various block sizes. We observe that P\&G is robust to variations in block size, achieving a similar average score across the five tasks regardless of the chosen block size. Furthermore, the sparse models maintain accuracy comparable to that of the dense baseline. For instance, on the WNLI dataset, P\&G achieves performance on par with the dense model across all settings. In all experiments, we iteratively sparsify the model during fine tuning to recover the accuracy of the dense version (see \Cref{sec:blast}). 

\begin{table}
\resizebox{\columnwidth}{!}{%
  \begin{tabular}{ccccccc}
    \toprule
    \multicolumn{2}{c}{\textbf{\png{} (\ref{sec:blocked-prune-and-grow})}} & \textbf{SST-2} & \textbf{MRPC} & \textbf{RTE} & \textbf{WNLI} & \textbf{Avg.}\\
    \textbf{$s_{max}$} & $b$ & ACC & ACC / F1 & ACC & ACC & \textbf{Score}\\
    \midrule
    70\% & 32  & 84.98 & 72.06/80.94 & 59.57 & 56.34 & 69.97 \\
    80\% & 32  & 85.89 & 68.14/77.35 & 61.01 & 56.34 & 68.90 \\
    90\% & 32  & 84.63 & 70.34/79.17 & 61.37 & 56.34 & 69.19 \\
    95\% & 32  & 85.32 & 70.83/79.73 & 59.21 & 56.34 & 69.13 \\
    70\% & 64  & 84.40 & 66.42/77.28 & 59.21 & 56.34 & 67.77 \\
    80\% & 64  & 85.67 & 69.36/79.13 & 60.29 & 56.34 & 69.12 \\
    90\% & 64  & 84.29 & 68.87/77.36 & 58.12 & 56.34 & 67.75 \\
    95\% & 64  & 85.32 & 68.38/78.96 & 55.23 & 56.34 & 67.78 \\
    70\% & 128 & 86.24 & 70.83/79.86 & 60.65 & 56.34 & 69.84 \\
    80\% & 128 & 85.55 & 70.59/79.87 & 53.07 & 56.34 & 68.13 \\
    90\% & 128 & 85.09 & 69.12/80.25 & 59.21 & 56.34 & 69.02 \\
    95\% & 128 & 83.94 & 68.87/79.81 & 54.87 & 56.34 & 67.72 \\
    \midrule
    \multicolumn{2}{c}{\textbf{Dense}}   & 87.04 & 73.28/81.22 & 62.45 & 56.34 & 71.08\\
    \bottomrule 
  \end{tabular}%
}
\captionof{table}{Fine tuning Llama 3.2 1B with blocked prune and grow for GLUE benchmark. Blocked prune and grow is robust to the choice of tile size, capable of lossless sparsification up to 95\% for WNLI.}
\label{tab:glue-benchmark-llama32}
\end{table}

\Cref{tab:glue-benchmark-bert} presents the accuracy results for the BERT-335M\cite{Devlin2019} model on a subset of the GLUE benchmark using oLLM (\Cref{sec:ollm}). Although the average score for Llama 3.2-1B in \Cref{tab:glue-benchmark-llama32} using P\&G falls below the dense baseline, the sparsified BERT model occasionally surpasses dense performance. As observed in the text summarization task from \Cref{sec:summarization-task}, oLLM's improved curvature estimation yields better accuracy under blocked sparsity.

Collectively, these results demonstrate the robustness of the proposed sparsification algorithms to varying sparsity levels, and block sizes for reconstructing accuracy, making \method{} a ready to use method for pretrained networks when speedup and memory reduction are of importance.

\begin{table}
\resizebox{\columnwidth}{!}{%
  \begin{tabular}{ccccccc}
    \toprule
    \multicolumn{2}{c}{\textbf{oLLM (\ref{sec:ollm})}}  & \textbf{SST-2} & \textbf{MRPC} & \textbf{RTE} & \textbf{WNLI} & \textbf{Avg.}\\
    \textbf{$s_{max}$} & $b$ & ACC & ACC / F1 & ACC & ACC & \textbf{Score}\\
    \midrule
70\%                    & 1          & 90.71          & 68.38/81.22     & 55.95        & 56.33         & 67.66                             \\
80\%                    & 1          & 87.72                   & 68.38/81.24     & 55.95        & 56.33         & 66.66                             \\
90\%                    & 1          & 89.44          & 68.38/81.29     & 53.79        & 56.33         & 66.52                             \\
95\%                    & 1          & 88                      & 68.38/81.34     & 53.06        & 56.33         & 65.79                             \\
70\%                    & 16         & 88.64          & 68.38/81.22     & 53.79        & 56.33         & 66.25                             \\
80\%                    & 16         & 87.38                   & 68.38/81.25     & 55.95        & 56.33         & 66.55                             \\
90\%                    & 16         & 83.25                   & 68.38/81.30     & 53.42        & 56.33         & 64.33                             \\
95\%                    & 16         & 84.74                   & 68.38/81.35     & 53.42        & 56.33         & 64.83                             \\
70\%                    & 32         & 89.79          & 68.38/81.22     & 53.42        & 56.33         & 66.51                             \\
80\%                    & 32         & 84.28                   & 68.38/81.26     & 55.59        & 56.33         & 65.4                              \\
90\%                    & 32         & 83.37                   & 68.38/81.31     & 53.79        & 56.33         & 64.49                             \\
95\%                    & 32         & 85.8                    & 68.38/81.36     & 54.15        & 56.33         & 65.42                             \\
70\%                    & 64         & 89.9          & 68.38/81.22     & 53.79        & 56.33         & 66.67                             \\
80\%                    & 64         & 85.34                   & 68.38/81.27     & 56.67        & 56.33         & 66.11                             \\
90\%                    & 64         & 88.76          & 68.38/81.32     & 55.95        & 56.33         & 67.01                             \\
95\%                    & 64         & 87.5                    & 68.38/81.37     & 53.06        & 56.33         & 65.63                             \\
70\%                    & 128        & 90.82          & 68.38/81.23     & 53.79        & 56.33         & 66.98                             \\
80\%                    & 128        & 86.92                   & 68.38/81.28     & 54.15        & 56.33         & 65.8                              \\
90\%                    & 128        & 87.72                   & 68.38/81.33     & 53.06        & 56.33         & 65.70                             \\
95\%                    & 128        & 85.73                   & 68.38/81.38     & 54.87        & 56.33         & 65.64                             \\
    \midrule
    \multicolumn{2}{c}{\textbf{Dense}}  & 92.88          & 68.38/81.39     & 47.5         & 56.33         & 65.57 \\
    \bottomrule 
  \end{tabular}%
}
\captionof{table}{Performance of BERT-335M on GLUE when fine-tuned with the oLLM pruning method (\Cref{sec:ollm}). Across a broad range of block sizes, oLLM maintains accuracy close to, or above the dense model, illustrating that second-order, block-structured pruning can be applied without substantial degradation on standard language understanding tasks.}
\label{tab:glue-benchmark-bert}
\end{table}

\subsubsection{Image Recognition (CIFAR10)}
\label{sec:pretraining-vits}

We demonstrate that P\&G (\Cref{sec:blocked-prune-and-grow}) can be applied not only to NLP tasks but also to the vision domain. We fine-tune a pretrained ViT~\cite{dosovitskiy2020image} on CIFAR-10~\cite{krizhevsky2009learning}, using weights originally trained on ImageNet~\cite{deng2009imagenet}. \Cref{tab:vit_pretraining} shows a comparison of four different ViT architectures under increasing sparsity levels achieved by P\&G. These results demonstrate the robustness of \method{} in handling increased sparsity levels in the MLP blocks of the ViT architecture. 

\begin{table}[!htbp] 
\centering
\resizebox{\columnwidth}{!}{%
  \begin{tabular}{cccccc}
    \toprule
    \textbf{Model} & \textbf{Dense} & \textbf{\png{}-70\%} & \textbf{\png{}-80\%} & \textbf{\png{}-90\%} & \textbf{\png{}-95\%} \\
    \midrule
    ViT-B/16 & 98.57\% & 93.45\% & 93.82\% & 94.56\% & 93.08\% \\
    ViT-B/32 & 98.54\% & 92.91\% & 92.95\% & 92.70\% & 91.49\% \\
    ViT-L/16 & 98.88\% & 95.82\% & 95.71\% & 95.48\% & 95.31\% \\
    ViT-L/32 & 98.75\% & 92.13\% & 93.36\% & 93.59\% & 93.57\% \\
    \bottomrule
  \end{tabular}%
}
\caption{Accuracy of training four ViT architectures with \png{} (\ref{sec:blocked-prune-and-grow}). ViT is pretrained on a large vision task~\cite{dosovitskiy2020image}, demonstrating the knowledge transfer capability of this sparsification algorithm.}
\label{tab:vit_pretraining}
\end{table}

\begin{figure}
  \centering
  \includegraphics[width=0.9\linewidth]{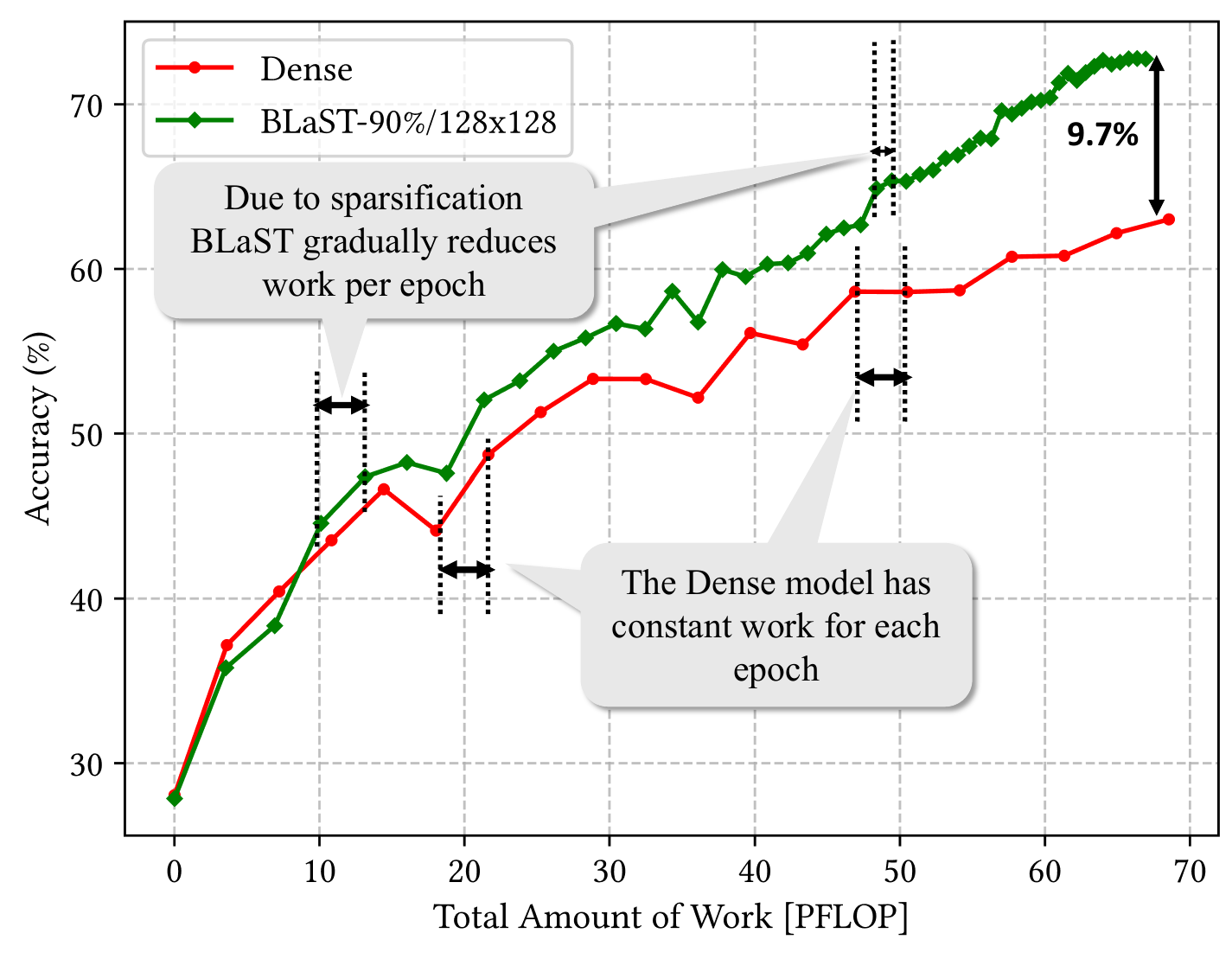}
  \caption{Training 90\%/128x128 for ViT-B/16 on CIFAR-10 with P\&G (\ref{sec:blocked-prune-and-grow}). Fewer PFLOP to process the entire training
set are required with sparsification as we sparsify the network. P\&G achieves higher accuracy with fewer FLOP, resulting in a better accuracy-to-FLOP ratio.}
  \label{fig:time_vs_flops}
\end{figure}

Furthermore, in \Cref{fig:time_vs_flops} we illustrate how test accuracy evolves for P\&G and the dense network as we iteratively train and sparsify the ViT architecture on CIFAR-10. We plot the test accuracy after every epoch (i.e., after each complete pass through the training data). We observe two key trends. First, as training progresses, P\&G requires fewer FLOP to process the entire training set. Second, P\&G achieves higher accuracy with fewer FLOP, resulting in a superior accuracy-to-FLOP ratio compared to the dense network.

\subsection{Memory Footprint for Inference}
\label{sec:memory-footprint-inference}

\begin{figure}
  \centering
  \includegraphics[width=0.9\linewidth]{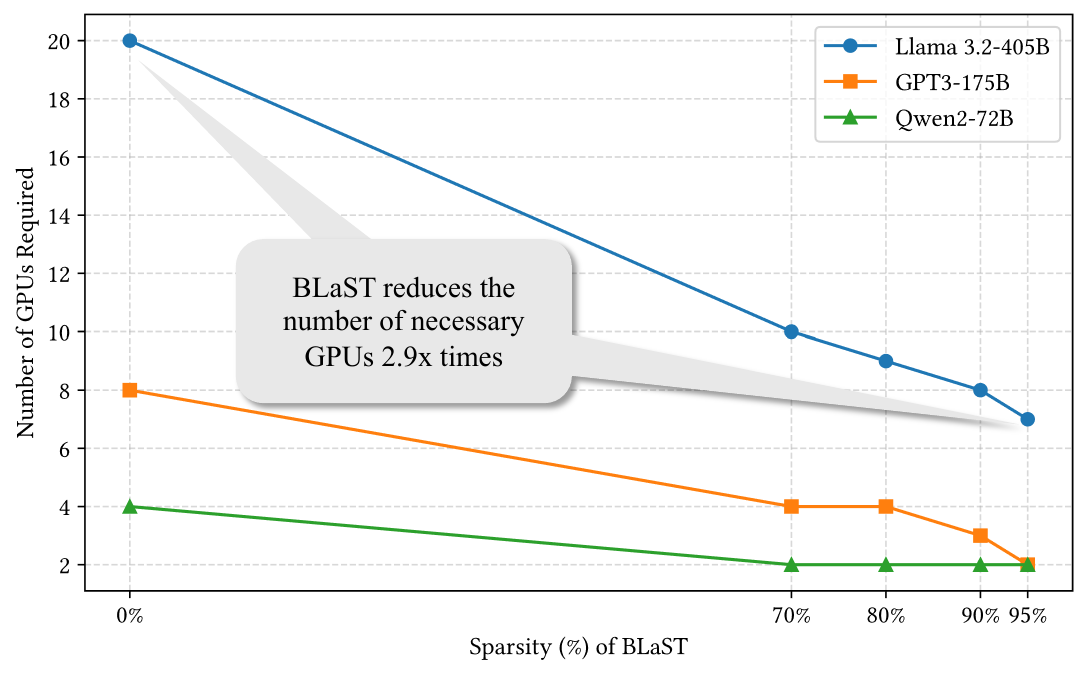}
  \caption{Number of GH200 GPUs required for storing weights in FP32 of the given model assuming 96GB per GPU. \method{} leads up to 4.45x inference memory reduction footprint.}
  \label{fig:memory-footprint}
\end{figure}

Pruning weights by up to 95\% significantly reduces the total parameter count and correspondingly decreases the model's memory footprint. \Cref{fig:memory-footprint} shows the number of NVIDIA GH200 GPUs required to store model weights, assuming 96GB HBM per GPU and FP32 precision. \method{} reduces the GPU requirements to approximately one-third of those needed for the dense Llama 3.2 405B model. This reduction directly translates to lower capital investments for inference data centers and decreased operational energy consumption.

\subsection{\method{} for Pretraining}
\label{sec:results-pretraining}

\begin{figure}
  \centering
  \includegraphics[width=\linewidth]{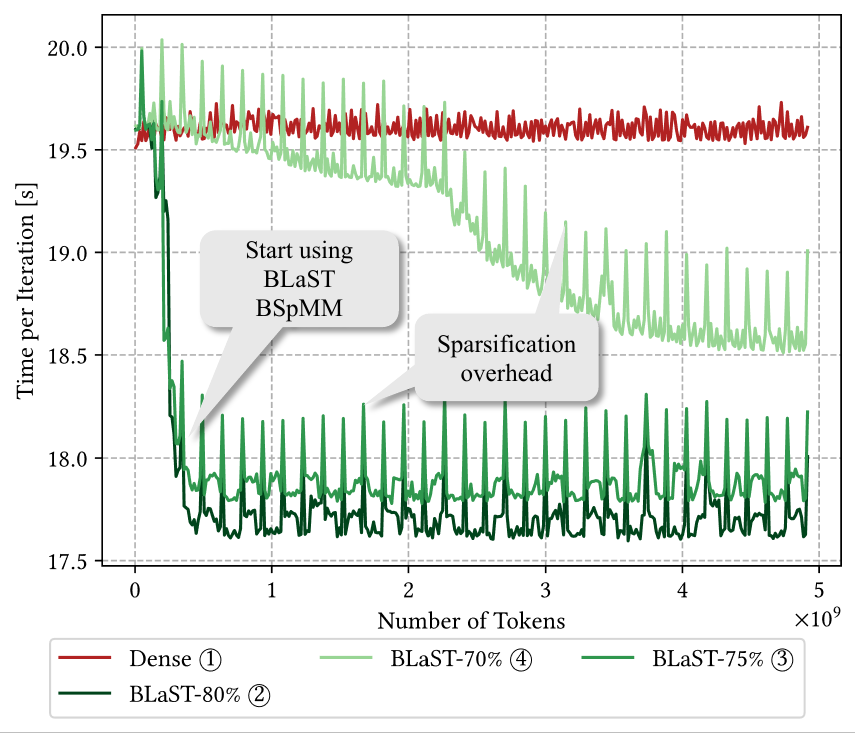}
  \caption{GPT2-XL pretraining with 4.9B tokens.}
  \label{fig:gpt2-performance-gh200}
  \caption{Time per iteration for GPT2-XL with OpenWebText~\cite{Gokaslan2019}. Each line plot uses the hyper parameters from \Cref{tab:perf-results}.}
\end{figure}

\subsubsection{Causal Language Modelling}
We pretrain GPT2-XL on the OpenWebText~\cite{Gokaslan2019} dataset with Causal Language modelling (CLM). The dataset consists of 8,012,769 documents (9,035,582,198 tokens), out of which 4,007 documents are used for building the test dataset. The model is trained on 4,915,200,000 tokens ($\approx$4.9B). \Cref{sec:perf-of-pre-training} shows the performance and accuracy for pretraining GPT2 and Llama 3.2 models. We use only the P\&G algorithm for pretraining due to the higher computational cost of oLLM, as shown in \Cref{sec:ollm}. 

\subsubsection{Accuracy and Performance of Pretraining}
\label{sec:perf-of-pre-training}

\Cref{tab:perf-results} shows the end-to-end time and perplexity for GPT2-XL and Llama 3.2 models with \method{} configurations. The GPT2-XL model pretraining shows a maximum speedup of 10.3\% using 80\% maximum sparsity (\textcircled{2}) compared to the dense model (\textcircled{1}). 
\begin{table}
\resizebox{\columnwidth}{!}{%
  \begin{tblr}{
    cells = {c},
    cell{2}{1} = {r=3}{},
    cell{2}{9} = {r=3}{},
    cell{2}{10} = {r=3}{},
    cell{5}{1} = {r=2}{},
    cell{5}{9} = {r=2}{},
    cell{5}{10} = {r=2}{},
    hline{1-2,5} = {-}{},
  }
  \textbf{Model} & \textbf{$b$} & \textbf{$s_{max}$} & $step\_size$ & \textbf{$d$} & $L$ & {\textbf{\method{}}\\\textbf{(h)}} & {\textbf{\method{}}\\\textbf{PPL}} & {\textbf{Dense}\\\textbf{(h)}} & {\textbf{Dense}\\\textbf{PPL}} \\
   GPT2-XL        & 128                 & 80\%                 & 100                                           & 9000                 & 2                                                     & 49.36  \textcircled{2} & 5.19                            & 54.45 \textcircled{1}                         & 4.79                           \\
                 & 128                 & 75\%                 & 100                                           & 9000                 & 2                                                     & 49.86  \textcircled{3} & 5.14                            &                                &                                \\
                 & 64                  & 70\%                 & 100                                           & 0                    & 2                                                     & 53.00 \textcircled{4} & 5.11                            &                                &                                \\                            
  \end{tblr}
}
\captionof{table}{Pretraining results for P\&G (\ref{sec:blocked-prune-and-grow}) with GPT2-XL on OpenWebText. \method{} reduces end-to-end time of while preserving accuracy.}
\label{tab:perf-results}
\vspace{-10pt}
\end{table}
\Cref{fig:gpt2-performance-gh200} shows the time per iteration over 4.9 billion tokens for the GPT2-XL model. The performance is compared for various sparsity configurations of \method{} against the dense case. The spikes observed for the sparse configurations are caused by mask generation, which occurs every 100 iterations. Dense matrix multiplication is used until the sparsity reaches 60\%. At 60\% sparsity, the BSpMM routines (\Cref{sec:kernel_section}) take over, which leads to a sharp drop in per-iteration time as sparsity increases.
\method{}-80\%/128x128 (plot \textcircled{2}) and \method{}-75\%/128x128 (plot \textcircled{3}) configurations exhibit an earlier activation of SpMM routines due to higher values of $d$.

The previous results demonstrate that \method{} with P\&G can be used to speed up the pretraining of Transformer based architectures with negligible loss of accuracy.

\section{Related Work}
\label{sec:related}
\paragraph{\textbf{Efficient ML}}
Various innovations, both algorithmic and hardware-based, have been proposed to reduce the time and cost associated with training and inference in large neural networks. These methods encompass quantization~\cite{nagel2021white,ashkboos2024quarot,dettmers2023spqr,frantar2023gptqaccurateposttrainingquantization,ashkboos2023quik}, efficient dataset selection~\cite{okanovicrepeated,bartoldson2023compute,mirzasoleiman2020coresetsdataefficienttrainingmachine,yang2023towards,killamsetty2021grad,killamsetty2021glister}, and weight pruning~\cite{Hoefler2021}. In addition, tailored AI chips and GPUs with mixed precision compute units~\cite{Sean2023,Jouppi2017,Luo2024,Lee2024a,Abts2022,Markidis2018,Schieffer2024,Kalamkar2019,li2022efficient,Okanovic2024} have been designed to efficiently exploit memory bandwidth and parallelism, thereby accelerating matrix multiplications.

Dynamic low-rank approximation (DLRA) techniques~\cite{Koch2007,Schotthofer2022,Ceruti2022,Ceruti2022a} compress CNNs by leveraging rapidly decaying singular values of weight matrices. However, in transformer-based models these singular values decay slowly~\cite{Schotthofer2024,Cheng2005}, reducing DLRA’s applicability. In contrast, weight pruning aims to remove individual weights, and has been successfully applied in both pre-training and post-training settings~\cite{Hu2024,Hubara2021,Zhou2021,Zhou2021a,Evci2021,Frankle2019,Chen2020,Child2019,Pan2024,Frantar2023,Ashkboos2024,Ivanov2023,Hu2021,Buyukakyuz2024,Chen2024}. Structured sparsity approaches, such as 2:4 sparsity~\cite{Hu2024a,Mishra2021}, coexist with unstructured methods based on the lottery ticket hypothesis~\cite{Frankle2019,Chen2021,You2022}. Additional strategies include the use of straight-through estimators~\cite{Hu2024a} and transposable masks with minimum variance estimators~\cite{Hubara2021,Chmiel2024}. Moreover, dynamic 'prune and grow' methods that exploit first-order gradient momentum~\cite{Evci2021,Abdelfattah2024,cusparse,hipsparse} and knowledge distillation from large teacher networks~\cite{gou2021knowledge,hinton2015distillingknowledgeneuralnetwork,muralidharan2024compact} further enhance sparsity.

\paragraph{\textbf{Matrix Multiplication}}
Sparse matrix multiplication on parallel architectures remains an active research area. Surveys summarize many effective techniques~\cite{gao2023,FilipponeReview}, while recent work exploits dense MMA units to accelerate sparse computations~\cite{lu2023,niu2020,Zachariadis2020,Okanovic2024}. For instance, DASP~\cite{lu2023} employs MMA units for SpMV acceleration, further boosted by \CLUB, which exposes dense substructures in sparse matrices.
Reordering methods are key to improving data locality, bandwidth, and load balancing in sparse kernels~\cite{peng2020,Pichel2005,Pinar99,spMMHong,gianinazzi2024arrow,Ma2025b}. Graph-partitioning has boosted SpMV on multicore and GPU platforms~\cite{Trotter2023,Pichel2012,pichon2017,silva2017influence} using techniques such as Reverse Cuthill–McKee~\cite{RMC}, approximate minimum degree~\cite{Amestoy2004}, nested dissection~\cite{karypis1998fast}, and Gray code ordering~\cite{Zhao2020}. Recent advances, including community-based and just-in-time parallel reordering (Rabbit Order)~\cite{Balaji2023,rabbit} and others~\cite{gleinig2022optimal}, further enhance cache locality and performance.
Similarity-based reordering has been integrated into novel SpMM kernel designs. Hierarchical clustering via locality-sensitive hashing supports adaptive sparse tiling~\cite{peng2020,spMMHong}, while similarity-based clustering improves both variable-size block multiplication~\cite{sylos2022}, and the use of matrix units in block-based SpMM kernels~\cite{Okanovic2024}.

\section{Conclusion}
This paper introduces \method{}, a general, scalable, and widely applicable end-to-end framework that greatly reduces memory footprint, hardware cost of running contemporary large-scale LLMs, while simultaneously reducing runtime in all scenarios, both training- and inference-related.
\method{} iteratively sparsifies the weights of the linear layers inside the Transformer blocks, achieving up to 95\% sparsity with negligible accuracy loss. 
Beyond its current implementation, \method{} provides a flexible foundation that can seamlessly integrate with any future block pruning algorithm, ensuring its continued relevance as sparsification techniques evolve. 
\method{} introduces a highly efficient BSpMM kernel that can be used as a standalone sparse linear algebra primitive, beyond ML workloads.
We perform a comprehensive performance study of the proposed sparse kernel on state-of-the-art neural network architectures.
The kernel delivers up to $521\times$ speedup over state of the art SpMM, up to $23\times$ over dense GEMM, $1.6\times$ end to end speedups, and $2.2\times$ distributed inference speedups on the Llama3 family.
These gains hold across a range of block sizes, which provide a practical knob to trade speed for accuracy. We see \method{} as a reliable building block for efficient inference and training pipelines at large scale.

\bibliographystyle{IEEEtran}
\bibliography{bibliography}
\end{document}